%% file: paper.tex
\documentclass[letterpaper, 10 pt, conference]{ieeeconf}  

\IEEEoverridecommandlockouts                              

\title{\LARGE \bf
TrackVLA++: Unleashing Reasoning and Memory Capabilities in VLA Models for Embodied Visual Tracking
}

\input{package}

\input{macro}

\begin{document}

\author{%
    \begin{minipage}{\linewidth}
    \centering
    \vspace{8pt}
	Jiahang Liu$^{1,2,*}$ ~
	Yunpeng Qi$^{3,4,*}$ ~
    Jiazhao Zhang$^{1,2,*}$ \\
	Minghan Li$^{2}$ ~
    Shaoan Wang$^1$ ~
    Kui Wu$^5$ ~
    Hanjing Ye$^6$ \\
    Hong Zhang$^6$ ~
    Zhibo Chen$^3$ ~
    Fangwei Zhong$^7$~
    Zhizheng Zhang$^{2,4,\dagger} ~$
    He Wang$^{1,2,4,\dagger}$ 
    \end{minipage}
	\\ 
    \begin{minipage}{\linewidth}
    \centering
    \begin{tabular}{c}
    \normalsize{$^1$Peking University~
	$^2$Galbot~
    $^3$USTC~
    $^4$BAAI} \\
    \normalsize{$^5$Beihang University~
    $^6$SUSTech ~
    $^7$Beijing Normal University}
    \end{tabular}
    \end{minipage}
    \\
    \begin{minipage}{\linewidth}
    \centering
    \begin{tabular}{c}
    Project Page: \url{https://pku-epic.github.io/TrackVLA-plus-plus-Web/}
    \end{tabular}
    \end{minipage}
\thanks{
$^*$ Equal Contribution, $^{\dagger}$ Equal Advising
}
}

\maketitle

\begin{strip}
  \centering
  \vspace{-3em} 
  \includegraphics[width=\textwidth]{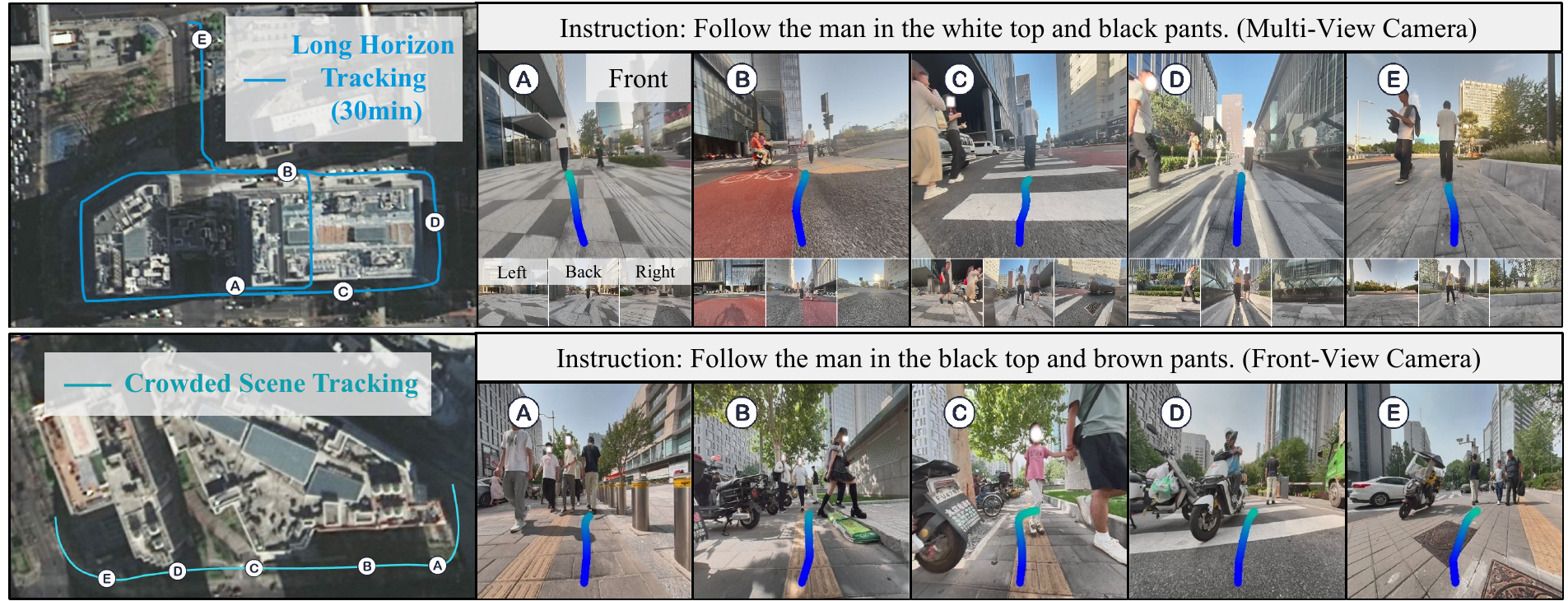}
  \captionof{figure}{\textbf{Real-world demonstration of \ours}. 
  \ours is a novel Vision-Language-Action model that incorporates spatial reasoning and target identification memory, enabling superior performance in both long-horizon and highly crowded tracking scenarios.
  }
  \label{fig:teaser}
  \vspace{-0.5em}
\end{strip}

\thispagestyle{empty}
\pagestyle{empty}

\input{section/abstract}

\input{section/intro}

\input{section/related_works}

\input{section/overview}

\input{section/architecture}

\input{section/experiments}

\input{section/conclusion}

\bibliographystyle{IEEEtran}
\bibliography{IEEEabrv,references}


\end{document}

%% file: package.tex
\usepackage[table]{xcolor}
\usepackage{titletoc}
\usepackage{tocloft}
\usepackage{lipsum}
\usepackage{url}
\usepackage{xspace}
\usepackage{graphicx}
\usepackage{subcaption}
\usepackage{listings}
\usepackage[nolist]{acronym}
\usepackage{amsmath}
\usepackage[font={small}]{caption}
\usepackage[export]{adjustbox}
\usepackage{amssymb}
\usepackage{bm}
\usepackage{booktabs}
\usepackage{balance}
\usepackage{cite}
\usepackage{color}
\usepackage{dsfont}
\usepackage{esvect}
\usepackage{float}
\usepackage{graphicx}
\usepackage{import}
\usepackage{mathtools}
\usepackage{multirow}
\usepackage{multicol}
\usepackage{flushend}
\usepackage{outline}
\usepackage{paralist}
\usepackage{siunitx}
\usepackage{stackengine}
\usepackage{stmaryrd}
\usepackage{systeme}
\usepackage{soul}
\usepackage{url}
\usepackage{tikz}
\usepackage{xcolor}
\usetikzlibrary{tikzmark}
\usetikzlibrary{calc}
\usepackage[normalem]{ulem}
\let\labelindent\relax
\usepackage{enumitem}
\usepackage{algorithm2e}
\usepackage{scalerel}
\usepackage{makecell}
\usepackage{graphicx}
\usepackage{stfloats}
\usepackage{graphicx}
\usepackage{caption}
\usepackage{graphicx}
\usepackage{caption}
\usepackage{cuted}

\definecolor{citecolor}{HTML}{0071bc}
\usepackage[pagebackref=true,breaklinks=true,urlcolor=blue,colorlinks,citecolor=citecolor,bookmarks=false]{hyperref}

%% file: macro.tex
\def\ours{TrackVLA++\xspace}

\newcommand{\eg}{\textit{e.g.,}\xspace}

\definecolor{myblue}{RGB}{0,100,200}
\definecolor{mygreen}{HTML}{009900}

\definecolor{mypurple}{HTML}{CDA0CD}

\definecolor{BestColor}{HTML}{ffffff}
\definecolor{SecondColor}{HTML}{ffffff}

%% file: section/abstract.tex
\begin{abstract}
Embodied Visual Tracking (EVT) is a fundamental ability that underpins practical applications, such as companion robots, guidance robots and service assistants, where continuously following moving targets is essential.
Recent advances have enabled language-guided tracking in complex and unstructured scenes. However, existing approaches lack explicit spatial reasoning and effective temporal memory, causing failures under severe occlusions or in the presence of similar-looking distractors.
To address these challenges, we present \ours, a novel Vision–Language–Action (VLA) model that enhances embodied visual tracking with two key modules: a spatial reasoning mechanism and a Target Identification Memory (TIM). The reasoning module introduces a Chain-of-Thought paradigm, termed Polar-CoT, which infers the target’s relative position and encodes it as a compact polar-coordinate token for action prediction. Guided by these spatial priors, the TIM employs a gated update strategy to preserve long-horizon target memory, ensuring spatiotemporal consistency and mitigating target loss during extended occlusions.
Extensive experiments show that \ours achieves state-of-the-art performance on public benchmarks across both egocentric and multi-camera settings.
On the challenging EVT-Bench DT split, \ours surpasses the previous leading approach by 5.1$\%$ and 12$\%$ respectively. Furthermore, \ours exhibits strong zero-shot generalization, enabling robust real-world tracking in dynamic and occluded scenarios.
\end{abstract}

%% file: section/intro.tex
\section{Introduction}
Embodied Visual Tracking (EVT) is a fundamental yet challenging task, where an agent navigates in dynamic physical environments and continuously track a specified moving target based on visual perception.
Recent methods have shown remarkable progress in this task~\cite{maalouf2024follow,zhang2018coarse,zhong2019ad,zhong2023rspt,li2020pose,zhong2024empowering}. 
Recent advancements in EVT increasingly leverage the powerful generalization capability of pre-trained Visual Foundation Models (VFMs)~\cite{kirillov2023segment, ravi2024sam, liu2023grounding} to enhance target identification from visual inputs.
Building on this perceptual foundation, agents employ policy learning techniques, such as imitation learning\cite{zhang2024uni} or reinforcement learning~\cite{zhong2019ad, luo2018end,zhong2024empowering}, to generate actions that enable effective target pursuit.

More recently, leveraging large language models (LLMs) has introduced a promising new paradigm for the EVT task. Pioneering works, notably TrackVLA~\cite{wang2025trackvla} and LOVON~\cite{peng2025lovon}, exemplify this trend by integrating powerful Vision-Language Models (VLMs) to handle complex, language-guided tracking tasks.
TrackVLA, for instance, introduces a unified, end-to-end Vision-Language-Action (VLA) framework that learns a holistic tracking policy. 
It processes visual-language inputs using a VLM, with the latent representations decoded into tracking trajectories through an anchor-based diffusion policy.
This design not only demonstrates strong sim-to-real generalization and real-time performance but also benefits from the tight coupling of perception and planning, which effectively mitigates the information loss and error propagation inherent in decoupled pipelines.
In contrast, LOVON adopts a hierarchical strategy, using LLM as a high-level planner to decompose instructions into simpler sub-tasks, which are then executed by a low-level motion model to predict immediate tracking actions.
Despite their advancements, these state-of-the-art (SOTA) methods lack explicit reasoning capability and robust mechanism for long-horizon target identification. As a result, their performance degrades in complex and unstructured scenes, particularly those involving severe occlusions or multiple visually similar distractors.

To address these challenges, we propose \ours, a novel VLA framework for the EVT task that is empowered with explicit spatial reasoning capability and effective temporal memory
to enable long-horizon target identification. At the core of our approach is the Polar Chain-of-Thought (Polar-CoT) mechanism, which enables spatial reasoning by inferring the target’s relative position, expressed as angle and distance in agent-centric polar coordinate system.
In contrast to prior CoT mechanisms in robot manipulation, which generate verbose textual plans or auxiliary visual intermediates (\eg bounding boxes or subgoal images)~\cite{deng2025graspvla,zhang2025inspire,zawalski2024ECoT,zhao2025cot-llava}, our Polar-CoT introduces a compact design that maintains inference efficiency by predicting only \textbf{one} reasoning token, which serves as the basis for the Target Identification Memory (TIM) module.
TIM is specifically designed to preserve a persistent and robust representation of the target’s visual identity over long horizons, even under challenging conditions such as prolonged occlusions. To this end, TIM employs a confidence-aware gating mechanism that strictly regulates memory updates: the memory state is refreshed only when Polar-CoT predicts the target’s presence with high confidence. During each update, TIM integrates its historical state with newly extracted visual features from the region specified by Polar-CoT’s spatial prediction, where the contribution of new observations is weighted in proportion to the confidence score. Furthermore, all the aforementioned techniques naturally extend to multi-view settings, where they not only retain compatibility but also deliver enhanced tracking performance.

We conducted extensive experiments to evaluate the effectiveness and generalization ability of \ours across both simulated benchmarks and real-world scenarios. Our method achieves SOTA performance in both egocentric and multi-camera settings. Specifically, on the highly challenging EVT-Benchmark~\cite{wang2025trackvla} \verb|DT split|, \ours outperforms previous leading methods by 5.1\% and 12\% in success rate for egocentric and multi-camera settings, respectively. Additionally, \ours accomplishes new SOTA results on the Gym-UnrealCV benchmark~\cite{qiu2017unrealcv}, which further demonstrates its superiority over existing methods. Beyond these benchmarks, \ours exhibits remarkable zero-shot generalization, demonstrating robust performance in real-world environments, as highlighted in Fig.~\ref{fig:teaser}, Fig.~\ref{fig:real_gallery} and our supplementary video. 
The contributions of this work can be summarized as follows:

\begin{itemize}
\item We propose a novel Polar-CoT mechanism for the EVT task, which equips the model with explicit spatial reasoning capability, achieving significant performance improvements while maintaining computational efficiency.
\item We propose the Target Identification Memory (TIM), a robust module for long-horizon target identification that leverages reasoning guided memory update to achieve resilience against severe occlusions and distractors.
\item We conduct extensive evaluations, showing that \ours achieves state-of-the-art performance across multiple simulation benchmarks and demonstrates remarkable generalization to real-world scenarios.
\end{itemize}

%% file: section/related_works.tex
\section{Related Works}
\noindent
\textbf{Vision-Language-Action Models.}
The paradigm of extending pre-trained Vision-Language Models (VLMs)\cite{shen2024longvu, steiner2024paligemma, chiang2023vicuna} with action-generation capabilities has established Vision-Language-Action (VLA) models as a cornerstone of modern embodied AI. This approach has yielded significant success in manipulation~\cite{black2024pi_0, intelligence2025pi_, kim2024openvla, qu2025spatialvla, zhong2025dexgraspvla, deng2025graspvla} and navigation~\cite{zhang2024uni, zhang2024navid, cheng2024navila}. Recently, the VLA paradigm was extended to the dynamic task of Embodied Visual Tracking (EVT), with models like TrackVLA~\cite{wang2025trackvla} achieving impressive results. In this work, we propose \ours, which enhances its predecessor with reasoning ability and long-horizon memory.

\noindent
\textbf{Embodied Visual Tracking (EVT)}~\cite{ye2024person, ye2025rpf, francis2025principles} requires an agent to continuously pursue a dynamic target based on its visual observations, relying on accurate target recognition and optimal trajectory planning. 
Early works~\cite{puig2023habitat,luo2018end,luo2019end,devo2021enhancing,zeng2024poliformer,zhong2021towards,bajcsy2024learning,scofano2024following,zhong2024empowering,shah2022offline} adopted a decoupled paradigm, pairing visual foundation models~\cite{kirillov2023segment} for perception with reinforcement learning for planning. 
Recently, the field has shifted towards end-to-end VLA models to support natural language inputs~\cite{zhang2024uni,wang2025trackvla,peng2025lovon}. Uni-NaVid~\cite{zhang2024uni} pioneered this direction with large-scale imitation learning, though its discrete action space limited real-world adaptability. 
Building on this, TrackVLA~\cite{wang2025trackvla} made significant advances by integrating recognition and planning into unified frameworks, showing strong performance in real-world tracking tasks. Similarly, LOVON~\cite{peng2025lovon} employs a hierarchical approach, where a high-level LLM planner breaks complex instructions into simpler sub-goals, executed by a low-level controller for navigation and tracking. Despite their success, both models still lack explicit reasoning capabilities and robust long-horizon target identification. In this work, we introduce \ours, a novel framework that enhances embodied visual tracking by incorporating a reasoning module and target identification memory.

\noindent
\textbf{Chain-of-Thought Reasoning for Embodied AI.} 
Chain-of-Thought (CoT) reasoning, which prompts models to think step-by-step, has proven effective for complex tasks~\cite{wei2022chain} and is increasingly adopted in VLA models to enhance reasoning and generalization ability~\cite{deng2025graspvla,zhang2025inspire,zawalski2024ECoT,mu2023embodiedgpt,cao2024cognav,zhang20233d}. 
A common strategy in these works, primarily focusing on robotic manipulation, is to generate explicit and computationally intensive intermediate representations (\eg such as high-level plans, object coordinates, or subgoal images) as prerequisites for final actions.
These can include high-level textual plans, object bounding boxes, grasping coordinates, subgoal images, or coarse-grained discrete directions~\cite{deng2025graspvla, zhao2025cot-llava, zawalski2024ECoT, zhang2025inspire}. While effective for manipulation tasks, these approaches can introduce significant inference overhead, making them unsuitable for highly dynamic scenarios like EVT.
In contrast, our method introduces an efficient CoT process especially designed to satisfy the dynamic demands of EVT, achieving robust reasoning while maintaining high inference efficiency.

%% file: section/overview.tex
\input{figure/pipeline}
\section{Overview}
\noindent
\textbf{Task Formulation.}
The task of Embodied Visual Tracking (EVT) can be formulated as: At each time step $T$, given a language description $\mathcal{L}$ of the target object and a set of on-the-fly captured RGB observations $\{\mathcal{O}_{T}^{N} \mid t=1,\dots,T,\; n=1,\dots,N \}$ from $N$ cameras, the agent is required to predict a continuous tracking trajectory $\mathcal{W}_T = \{ w_{1}, w_{2}, \dots \}$. Each waypoint $w_i = (x, y, \theta) \in \mathbb{R}^3$ defines a target displacement $(x, y)$ and a heading change $\theta$ within the agent's egocentric coordinate. 
The task is deemed successful if the agent maintains a predefined following distance $D$ from the target.

\medskip

\noindent
\textbf{Model Overview.} 
As shown in Fig.~\ref{fig:pipeline}, \ours is an end-to-end VLA model built upon the navigation foundation model NavFoM~\cite{zhang2025NavFoM}. To enhance tracking intelligence, \ours introduces two key improvements: a CoT-based spatial reasoning mechanism Polar-CoT and a long-horizon Target Identification Memory (TIM). 
Given an online-captured video stream, \ours extracts visual features from historical and current observations and predicts the reasoning token through the proposed Polar-CoT mechanism. Based on this prediction, the TIM tokens are adaptively updated to maintain a robust representation of the target's identity over time.
The reasoning token, updated TIM tokens, visual tokens and language tokens are then concatenated to form the input sequence for the large language model (implemented with Qwen2-7B~\cite{bai2025qwen2}). Leveraging this comprehensive context, the model predicts an action token, which is finally decoded by a MLP-based action head to predict the tracking trajectory.

%% file: figure/pipeline.tex
\begin{figure*}[t]
  \centering
    \includegraphics[width=\linewidth]{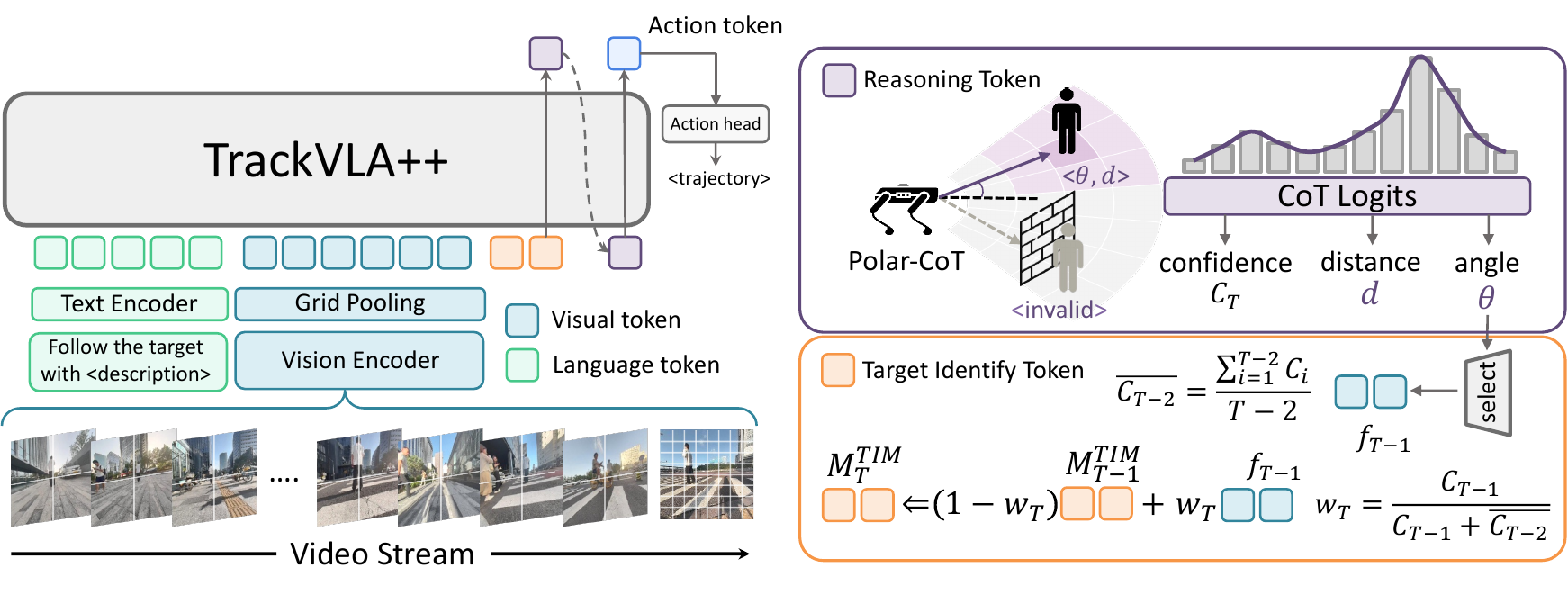}
  \caption{\textbf{The pipeline of \ours.} Given a video stream and a language instruction, \ours predicts a tracking trajectory by utilizing Polar-CoT reasoning to infer the target's position and continuously updating the Target Identification Memory with CoT-based predictions for long-horizon tracking.
  }
  \vspace{-1 em}
  \label{fig:pipeline}
\end{figure*}

%% file: section/architecture.tex
\section{Architecture}
\subsection{TrackVLA++ Architecture}
\noindent
\textbf{Observation Encoding.}
We process the on-the-fly video stream $\mathcal{O}_{1:T}^{1:N}$ by a dual-encoder architecture, extracting and concatenating visual features $\{V_{t}^{n} | t=1, ..., T, n=1, ..., N\}$ from SigLIP~\cite{zhai2023siglip} and DINOv2~\cite{oquab2023dinov2}. To mitigate the computational cost of long-horizon inputs, we then apply the grid pooling strategy~\cite{zhang2024navid}, generating a different resolution representation: $V^\text{fine} \in \mathbb{R}^{64 \times C}$, which consists of high-resolution features for the fine-grained details of the current observation and low-resolution features $V^\text{coarse} \in \mathbb{R}^{4 \times C}$ summarizing the coarse-grained historical context, where $C$ denotes the channel dimension.

To effectively manage the trade-off between long-range context and inference speed, our model employs a dual-memory architecture. For long-term memory, we introduce a fixed-size Target Identification Memory (TIM) to represent the target's history concisely. For short-term memory, we preserve the sliding window approach from TrackVLA, utilizing $k=32$ frames to form the current visual feature sequence, $V_T^\text{track}=\{V_{T-k}^\text{coarse}, \dots, V_{T-1}^\text{coarse}, V_T^\text{fine}\}$. The short-term visual sequence $V_T^\text{track}$ and the long-horizon TIM features $M_{T}^\text{TIM}$ are jointly projected into the LLM's latent space by a 2-layer MLP projector $\mathcal{P}(\cdot)$:
\begin{equation}
    E_T^V = \mathcal{P}(V_T^{track}), \quad E_T^M = \mathcal{P}(M_{T}^{TIM}),
\end{equation}

\noindent
\textbf{Polar-CoT Reasoning Forwarding.} 
To equip the model with spatial reasoning capability, we introduce a novel Polar Chain-of-Thought (Polar-CoT) mechanism, which is specifically designed for embodied visual tracking. In contrast to existing CoT approaches, which involve extensive reasoning steps, such as predicting object bounding boxes, Polar-CoT adopts a lightweight and agent-centric design based on polar coordinates. 
This CoT design stands in sharp contrast to traditional bounding box-based methods, which often suffer from computational inefficiency and ambiguity, particularly in multi-camera settings where overlapping fields of view (FoV) lead to redundant or conflicting predictions that are difficult to reconcile.

As demonstrated in Fig.~\ref{fig:pipeline}, Polar-CoT discretizes the agent’s perceivable annular FoV into a structured grid of sectors, where each sector is uniquely identified by a quantized combination of relative angle ($\theta$) and distance ($d$). This discrete combination is then encoded as a unique vocabulary token, forming a compact and unified spatial representation.
Moreover, this unified spatial representation inherently supports multi-camera setups by sidestepping the challenge of predicting bounding boxes, thereby eliminating ambiguity and ensuring consistent spatial reasoning across different views.

The reasoning process is structured as follows. First, the projected visual embeddings ($E_T^V$) and long-term memory embeddings ($E_{T}^M$) are concatenated with the language tokens ($E^L$) to form the input sequence for the LLM. The model then generates a reasoning token, $E_T^\text{CoT}$, which encodes the target’s spatial information (direction and proximity) into a compact representation. To further enhance robustness, the vocabulary is extended with a dedicated \verb|<invalid>| token, allowing the model to explicitly signal when the target is occluded or outside the agent's FoV. This reasoning process is formally defined as:
\begin{align}
    E_T^\text{CoT} = \text{LLM}(\text{Concat}[E_{T}^M, E_{T}^V, E^L]),
\end{align}

\noindent
\textbf{Reasoning Feedback Memory Update.}
\label{method:memory}
To maintain a stable target identity across occlusions, we introduce the Target Identification Memory (TIM), a confidence-gated mechanism that prevents memory corruption from distractors or drift during target absence. At each timestep $T$, the TIM state $M_T^\text{TIM}$ is updated from its previous state $M_{T-1}^\text{TIM}$ via a weighted average with a new candidate feature $f_{T-1}$:
\begin{equation}
M_T^\text{TIM} = (1 - w_T) \cdot M_{T-1}^\text{TIM} + w_T \cdot f_{T-1},
\end{equation}
where the candidate feature $f_{T-1}$ represents the visual embedding from the predicted target view, identified from fine-grained features $V_{T-1}^\text{fine}$ by the reasoning token $E_{T-1}^\text{CoT}$. An \verb|<invalid>| token signifies that the target is occluded or absent.

The weight $w_T$ modulates the update based on prediction certainty. It is formulated by normalizing the confidence score $C_{T-1}$ against the historical average confidence $\overline{C_{T-2}}$:
\begin{equation}
w_T = \frac{C_{T-1}}{\overline{C_{T-2}} + C_{T-1}}, \quad \text{with} \quad \overline{C_{T-2}} = \frac{1}{T-2}\sum_{i=1}^{T-2} C_i,
\end{equation}
The confidence score $C_{T-1}$ itself quantifies the certainty of the reasoning token $E_{T-1}^\text{CoT}$ and is calculated using the normalized entropy of its logits $\mathbf{P}$:
\begin{equation}
C_{T-1} = 1 - \frac{H(\text{softmax}(\mathbf{P}))}{\log K},
\end{equation}
where $H(p) = -\sum p_i \log p_i$ is the entropy over the $K$-sized reasoning vocabulary. Consequently, a confident, one-hot-like distribution yields $C_{T-1} \approx 1$ and a larger weight $w_T$, while an uncertain distribution results in $C_{T-1} \approx 0$, effectively suppressing the memory update.

The TIM is initialized to a null state ($M_1^\text{TIM} = \emptyset$) and adopts the first valid feature $f_1$ as its state at $T=2$. Subsequently, the update process is governed by confidence: a high score ($C_{T-1} \to 1$) allows the memory to integrate the new feature $f_{T-1}$, whereas a low score ($C_{T-1} \to 0$) preserves the previous state $M_{T-1}^\text{TIM}$. Crucially, an \verb|<invalid>| token at timestep $t$ forces its confidence $C_t$ to zero. This freezes the memory during the next update at $T=t+1$, thereby preserving the last reliable representation until the target is confidently re-detected and ensuring robust long-term tracking.





\noindent
\textbf{Action Forwarding.} 
After generating the reasoning token $E_T^\text{CoT}$ and updating the TIM memory $M_{T}^\text{TIM}$, the model predicts an action token $E_T^\text{pred}$. $E_T^\text{pred}$ is then decoded by an MLP-based action head into a sequence of waypoints $\mathcal{W}_T$. The action prediction process is formally defined as:
\begin{align}
  E_T^\text{pred} = \text{LLM}(\text{Concat}[E_{T}^M, E_{T}^V, E^L,E_T^\text{CoT}]),
  \label{eq:traj}
\end{align}
\vspace{-1.5em}
\begin{align}
\mathcal{W}_T &= \text{ActionHead}(E_T^\text{pred}),
\end{align}
The overall training objective is defined as a weighted sum of three loss terms: the trajectory planning loss $\mathcal{L}_{\text{traj}}$, reasoning loss $\mathcal{L}_{\text{reason}}$, and vanilla text prediction loss $\mathcal{L}_{\text{text}}$:
\begin{equation}
\mathcal{L} = \mathcal{L}_{\text{traj}} + \alpha \mathcal{L}_{\text{reason}} + \beta \mathcal{L}_{\text{text}},
\end{equation}
where $\alpha$ and $\beta$ are balancing hyperparameters, empirically set to $0.2$ and $0.5$, respectively. $\mathcal{L}_{\text{traj}}$ is defined as the Mean Squared Error (MSE) between the predicted waypoints $\hat{w}_i$ and the ground truth waypoints $w_i^\text{gt}$:
\begin{equation}
\mathcal{L}_{\text{traj}} = \sum_{i=1}^{M} \text{MSE}(\hat{w}_i, w_i^\text{gt}),
\end{equation}
where $M$ denotes the number of waypoints to predict and $\hat{w}_i$ and $w_i^{\text{gt}}$ denote the predicted and ground truth trajectory waypoints, respectively.
$\mathcal{L}_{\text{reason}}$ is formulated as the log-likelihood term over the reasoning token $E_T^\text{CoT}$, conditioned on the concatenated inputs:
\begin{equation}
\mathcal{L}_{\text{reason}} = - \log \mathbf{P}(E_T^\text{CoT} \mid \text{Concat}[E_{T}^M, E_T^V, E^L]).
\end{equation}
In alignment with the established practices from VLM training~\cite{liu2023llava}, the model is trained for a single epoch on the combined dataset, as detailed in Sec.~\ref{sec:data}. 


\input{table/evt_bench}
\subsection{Dataset Construction}
\label{sec:data}
\noindent
\textbf{Polar-CoT Tracking Data Collection.}
We constructed a large-scale dataset comprising one million multi-view embodied visual tracking samples from the EVT-Bench~\cite{wang2025trackvla} training split, using the Habitat 3.0~\cite{puig2023habitat} simulator. Each tracking sample includes a multi-view RGB tracking history, a target description, Polar-CoT annotations, and the corresponding expert trajectory $\mathcal{W}_\text{gt}$.
To generate the Polar-CoT annotations, we recorded the target's relative angle ($\theta$) and distance ($d$) with respect to the robot at each timestep. Additionally, we extracted semantic masks for the target from all views. If the total number of pixels in the target mask was below a predefined threshold of 2,500 pixels, we classified the target as either occluded or too distant, assigning it a \verb|<invalid>| flag.
Furthermore, to enhance generalization, we introduced randomization into the camera parameters, including position, height and FoV. Simultaneously, we introduced randomization in camera views to enhance data diversity, ensuring that data from the front camera was consistently retained while randomly sampling data from other cameras for augmentation.

\noindent
\textbf{QA Data Organization.}
In line with the TrackVLA~\cite{wang2025trackvla}, we co-trained the model by balancing tracking data with question-answering (QA) data in a 1:1 ratio. This approach was designed to enhance the model’s open-world recognition capabilities. Specifically, we incorporated 294K person identification samples from SYNTH-PEDES~\cite{zuo2024plip}, 205K image-based QA samples, and 501K video-based QA samples from publicly available datasets~\cite{shen2024longvu,liu2023llava}.
In total, the QA data contributed one million samples, bringing the combined training dataset to two million samples. 
This comprehensive dataset enables the model to effectively integrate trajectory tracking and open-world recognition capability.

%% file: table/evt_bench.tex
\begin{table*}[t]
    \caption{
    \textbf{Performance on EVT-Bench.} 
    The evaluation metrics are defined as follows: \textbf{Success Rate (SR)}, the proportion of episodes that the agent ends correctly oriented within 1–3m of the target; \textbf{Tracking Rate (TR)}, the proportion of timesteps with successful target tracking; and \textbf{Collision Rate (CR)}, the proportion of episodes terminated due to collisions. 
    $\dag$: Uses GroundingDINO as the detector. 
    $\ddag$: Uses SoM~\cite{yang2023set} + GPT-4o~\cite{openai2024introducing} as the visual foundation model. 
    \textbf{Bold} and \underline{underline} denote the best and second-best results, respectively.}
    \centering
    \begin{tabular}{lccccccccc}
        \toprule
        \multirow{2}{*}{Methods}& \multicolumn{3}{c}{\textit{Single-Target Tracking (STT)}} & \multicolumn{3}{c}{\textit{Distracted Tracking (DT)}} & \multicolumn{3}{c}{\textit{Ambiguity Tracking (AT)}} \\
        \cmidrule(lr){2-4} \cmidrule(lr){5-7} \cmidrule(lr){8-10}
        & SR$\uparrow$ & TR$\uparrow$ & CR$\downarrow$ & SR$\uparrow$ & TR$\uparrow$ & CR$\downarrow$ & SR$\uparrow$ & TR$\uparrow$ & CR$\downarrow$ \\
        \midrule
        IBVS$\dag$~\cite{gupta2016novel} & 42.9 & 56.2 & 3.75 & 10.6 & 28.4 & 6.14 & 15.2 & 39.5 & 4.90 \\
        PoliFormer$\dag$~\cite{zeng2024poliformer} & 4.67 & 15.5 & 40.1 & 2.62 & 13.2 & 44.5 & 3.04 & 15.4 & 41.5 \\
        EVT~\cite{zhong2024empowering} & 24.4 & 39.1 & 42.5 & 3.23 & 11.2 & 47.9 & 17.4 & 21.1 & 45.6 \\
        EVT$\ddag$~\cite{zhong2024empowering} & 32.5 & 49.9 & 40.5 & 15.7 & 35.7 & 53.3 & 18.3 & 21.0 & 44.9 \\
        Uni-NaVid~\cite{zhang2024uni} & 25.7 & 39.5 & 41.9 & 11.3 & 27.4 & 43.5 & 8.26 & 28.6 & 43.7 \\
        TrackVLA~\cite{wang2025trackvla} & \underline{85.1} & 78.6 & \textbf{1.65} & 57.6 & 63.2 & \underline{5.80} & \underline{50.2} & \textbf{63.7} & \underline{17.1} \\
        NavFoM~\cite{zhang2025NavFoM} (Single view) & 85.0 & \underline{80.5} & - & \underline{61.4} & \underline{68.2} & - & - & - & - \\
        \rowcolor{mypurple!20}
        \textbf{Ours (single view)} & \textbf{86.0} & \textbf{81.0} & \underline{2.10} & \textbf{66.5} & \textbf{68.8} & \textbf{4.71} & \textbf{51.2} & \underline{63.4} & \textbf{15.9} \\
        \midrule
        NavFoM~\cite{zhang2025NavFoM} (Four views) & \underline{88.4} & \underline{80.7} & - & \underline{62.0} & \underline{67.9} & - & - & - & - \\
        \rowcolor{mypurple!20}
        \textbf{Ours(Four views)} & \textbf{90.9} & \textbf{82.7} & \textbf{1.50} & \textbf{74.0} & \textbf{73.7} & \textbf{3.51} & \textbf{55.9} & \textbf{63.8} & \textbf{15.1} \\
        \bottomrule
    \end{tabular}
    \vspace{-1em}
    \label{tab:evt-bench}
\end{table*}

%% file: section/experiments.tex
\section{Experiments}

In this section, we present a series of experiments designed to answer the following questions:
\begin{itemize}
\item How does \ours perform in comparison to SOTA EVT models?
\item What is the practical performance and robustness of \ours in challenging, real-world scenarios?
\item What are the individual contributions of our core components: the Polar-CoT mechanism and the TIM module, to the overall performance?
\end{itemize}

\subsection{Experiment Setups}
\noindent
\textbf{Benchmarks.}
We evaluate our method using the EVT-Bench~\cite{wang2025trackvla} and Gym-UnrealCV~\cite{qiu2017unrealcv} benchmarks. EVT-Bench is a comprehensive benchmark for embodied tracking in complex indoor scenes with lots of distractors, including visually identical appearances and ambiguous instructions. Gym-UnrealCV evaluation focuses on tracking in unseen, high-fidelity environments, providing a robust test for generalization. Additionally, we utilize the visual recognition benchmark from~\cite{wang2025trackvla} to evaluate fine-grained, zero-shot recognition accuracy and efficiency.


\noindent
\textbf{Metrics.}
To evaluate tracking performance, we use the standard evaluation metrics from Gym-UnrealCV~\cite{qiu2017unrealcv} and EVT-Bench, including success rate (SR), average episode length (EL), tracking rate (TR), and collision rate (CR). 

\input{figure/robot}
\noindent
\textbf{Implementation Details.} 
\ours is built upon NavFoM~\cite{zhang2025NavFoM}, with the Polar-CoT module discretizing the agent's perceivable space (an annular region between $0.6$m and $5.0$m) into 60 angular and 30 distance slices, each represented as a unique special token. The TIM state $M_t^{TIM}$ is encoded by 4 tokens, while the predicted trajectory $\mathcal{W}_t$ comprises 8 future waypoints. 
The model is trained on 8 NVIDIA H100 GPUs for about one day, resulting in a total of 192 GPU hours.
For deployment, as illustrated in Fig.~\ref{fig:robot}, \ours operates on a Unitree GO2 quadruped robot equipped with four SG3S11AFxK cameras for multi-view RGB streaming. The video stream is sent to a remote server with an NVIDIA RTX 4090 GPU for processing, where Polar-CoT tokens and trajectory waypoints are generated. 

\input{figure/sim_gallery}
\subsection{Simulation Benchmark Results}
\noindent
\textbf{Performance on EVT-Bench.}
As shown in Table~\ref{tab:evt-bench} and Fig.~\ref{fig:sim_gallery}, we first evaluate our method on the challenging EVT-Bench benchmark. \ours demonstrates substantial improvements over existing approaches across all three sub-tasks in both egocentric and multi-view camera settings, establishing a new SOTA. Notably, \ours achieves particularly strong gains in the most challenging categories. For example, on the \verb|DT| (Distracted Tracking) task, \ours improves the Success Rate (SR) to 74.0\%, representing a significant leap from the 62.0\% achieved by NavFoM.
The notable improvements in all metrics highlight the strengths of \ours in robust recognition, long-horizon following and effective collision avoidance. Importantly, despite NavFoM being trained on a massive dataset of 10 million trajectories, \ours achieves superior performance with significantly less training data, underscoring its data efficiency and advanced modular design.

\noindent
\textbf{Zero-shot performance on Gym-UnrealCV.}
Beyond EVT-Bench, we evaluate the model's generalization ability on the Gym-UnrealCV benchmark in a zero-shot manner, using a front-view camera for fair comparison. As shown in Table~\ref{tab:unrealcv_bench} and Fig.\ref{fig:sim_gallery}, \ours achieves SOTA performance across all sub-tasks. In the \verb|Single Target| and \verb|Unseen Objects| categories, our method, like TrackVLA, achieves the perfect scores (EL=500, SR=1.00), successfully tracking the target for the maximum episode duration. Crucially, in the more challenging \verb|Distractor| task, where the agent must differentiate the target from identical distractors, \ours outperforms the previous best method, TrackVLA, with a higher SR and longer EL.

\noindent
\textbf{Performance on Visual Recognition}. 
To further evaluate the fine-grained recognition ability of \ours, we compare it with SOTA VLMs and tracking VLAs~\cite{jiang2025referring, yang2023lisa++, openai2024introducing, wang2025trackvla} on a zero-shot human recognition task involving distinguishing between two unseen human images from the SYNTH-PEDES dataset. As shown in Table~\ref{tab:recognitio}, \ours achieves a SOTA accuracy of 87.5\%, outperforming strong baselines such as SoM + GPT-4o (82.4\%), TrackVLA (80.7\%), and NavFoM (84.0\%).

In terms of computational efficiency, \ours maintains an inference speed of 4.8 FPS, which is comparable to NavFoM (5.1 FPS) and approximately \textbf{48$\times$ faster} than GPT-based baselines (SoM + GPT-4o). Despite a slight decrease in speed due to the Polar-CoT module (4.8 FPS vs. 5.2 FPS without Polar-CoT), it delivers a notable improvement in recognition accuracy (87.5\% vs. 83.0\%). This demonstrates the effectiveness of the Polar-CoT module in enhancing the model's reasoning capabilities while maintaining a strong balance between accuracy and efficiency.

\input{table/unrealcv_bench}
\input{table/recognition_ability}
\input{figure/real_gallery}
\subsection{Real World Results}

We evaluated \ours in three challenging real-world scenarios, with quantitative results shown in Fig.~\ref{fig:real_gallery}:
(A) \textbf{Obstacle}: The target is temporarily occluded by large obstacles, testing the model's robustness to target disappearance and its ability to re-identify the target.
(B) \textbf{Winding Path}: The target follows a complex, winding trajectory, evaluating the tracking fidelity amidst continuous changes in direction.
(C) \textbf{Distractor}: 
The target is challenged by a human distractor, which serves to evaluate the model's robustness in recognition and the ability to recover from interference.

Across these tasks, \ours outperforms TrackVLA by 14\%, 7\%, and 17\% respectively, demonstrating substantially improved robustness in real-world conditions.

\subsection{Ablation Study}
\input{table/ablations}

We conduct an ablation study on the \verb|DT| split of EVT-Bench (four views) to investigate the effectiveness of the proposed modules, as summarized in Table~\ref{tab:ablations}. 
The performance gains are primarily attributed to the proposed modules. Specifically, the CoT module improves the SR by 6.0\%, while the TIM module (4 tokens) contributes an additional 2.8\%. These results highlight the complementary benefits of these components in enhancing tracking performance.
Furthermore, we investigate the effect of varying the number of TIM tokens. To our surprise, increasing the token number from 4 to 16 does not result in a noticeable performance improvement, suggesting that the model can achieve robust tracking with concise token representations. This finding emphasizes the efficiency of our design in maintaining high performance with minimal computational overhead.

%% file: figure/robot.tex
\begin{figure}[t]
  \centering
    \includegraphics[width=0.7\linewidth]{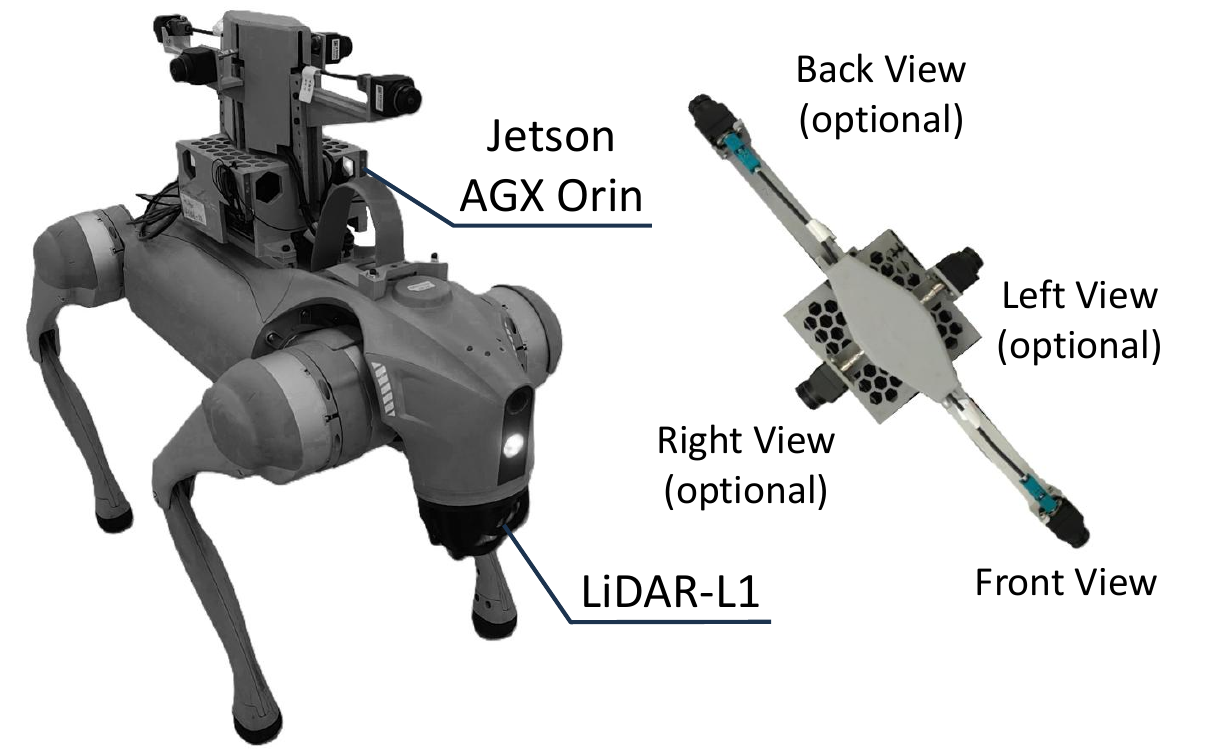}
  \caption{\textbf{Real-world system architecture.}}
  \label{fig:robot}
  \vspace{-1.5em}
\end{figure}

%% file: figure/sim_gallery.tex
\begin{figure}[t]
  \centering
    \includegraphics[width=0.9\linewidth]{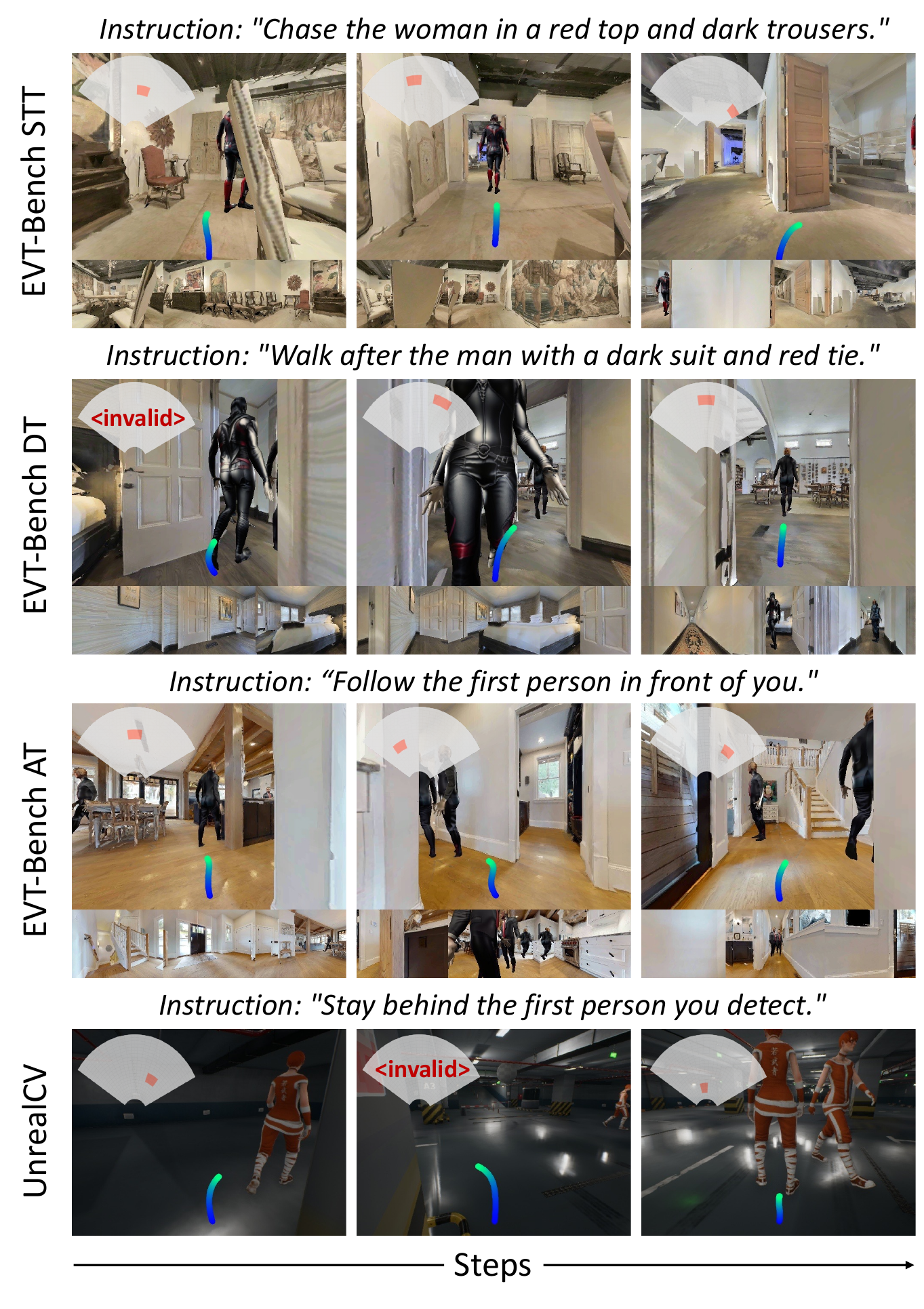}
  \caption{\textbf{Visualizations of the Simulation Experiments.} \ours performs well under occlusion and interference conditions. 
  The upper-left inset displays the Polar-CoT prediction, with the red area indicating the predicted target position, and the visualization  on EVT-Bench is cropped to a front sector for conciseness.
  Zoom in for a better view.
  }
  \label{fig:sim_gallery}
  \vspace{-1em}
\end{figure}

%% file: table/unrealcv_bench.tex
\begin{table}[t]
    \centering
    \caption{
    \textbf{Zero-shot Performance on Gym-UnrealCV.}  
    The evaluation metrics are defined as follows: \textbf{Episode Length (EL)}, the average number of steps before episode termination (maximum is 500); and \textbf{Success Rate (SR)}, the proportion of episodes completed for the full 500-step duration.
    $\dag$: \ours{} evaluated using only a single front-view camera for fair comparison.  
    \textbf{Bold} and \underline{underline} denote the best and second-best results, respectively.}
    \label{tab:unrealcv_bench}
    \begin{tabular}{lcccccc}
        \toprule
        \multirow{2}{*}{Methods} & \multicolumn{2}{c}{Single Target} & \multicolumn{2}{c}{Distractor} & \multicolumn{2}{c}{Unseen Objects} \\
        \cmidrule(lr){2-3} \cmidrule(lr){4-5} \cmidrule(lr){6-7}
        & EL$\uparrow$ & SR$\uparrow$ & EL$\uparrow$ & SR$\uparrow$ & EL$\uparrow$ & SR$\uparrow$ \\
        \midrule

        DiMP~\cite{bhat2019learning} & 367 & 0.58 & 309 & 0.27 & - & - \\
        SARL~\cite{luo2019end} & 394 & 0.57 & 240 & 0.14 & - & - \\
        AD-VAT~\cite{zhong2019ad} & 416 & 0.62 & 220 & 0.12 & - & - \\
        AD-VAT+~\cite{zhong2019ad+} & 454 & 0.76 & 224 & 0.12 & - & - \\
        TS~\cite{zhong2021towards} & 474 & 0.86 & 371 & 0.48 & - & - \\
        EVT~\cite{zhong2024empowering} & 490 & 0.95 & 459 & 0.81 & 480 & 0.96 \\
        TrackVLA~\cite{wang2025trackvla} & \textbf{500} & \textbf{1.00} & \underline{474} & \underline{0.91} & \textbf{500} & \textbf{1.00} \\
        \midrule
        \rowcolor{mypurple!20}
        Ours$^{\dag}$ & \textbf{500} & \textbf{1.00} & \textbf{484} & \textbf{0.92} & \textbf{500} & \textbf{1.00} \\
        \bottomrule
    \end{tabular}
    \vspace{-1em}
\end{table}

%% file: table/recognition_ability.tex
\begin{table}
    \centering
    \caption{
    \textbf{Comparison of Different Methods on Recognition Ability.}  
    $\dag$: Evaluation is restricted to the front-view setting for fair comparison.}
    \label{tab:recognitio}
    \begin{tabular}{lcc}
        \toprule
        Methods & ACC (\%) $\uparrow$ &  FPS $\uparrow$ \\
        \midrule
        RexSeek~\cite{jiang2025referring} & 54.3 & 1.1 \\
        LISA++~\cite{yang2023lisa++} & 78.2 & 0.6 \\
        SoM~\cite{yang2023set}+GPT-4o~\cite{openai2024introducing} & 82.4 & 0.1 \\
        TrackVLA~\cite{wang2025trackvla} & 80.7  & 10 \\
        NavFoM~\cite{zhang2025NavFoM} & \underline{84}  & 5.1 \\
        \midrule
        Ours$^{\dag}$ w/o Polar-CoT & 83  & 5.2 \\
        \rowcolor{mypurple!20}
        Ours$^{\dag}$ & \textbf{87.5}  & 4.8 \\
        \bottomrule
    \end{tabular}
    \vspace{-1.0em}
\end{table}

%% file: figure/real_gallery.tex
\begin{figure*}[t]
  \centering
    \includegraphics[width=1.0\linewidth]{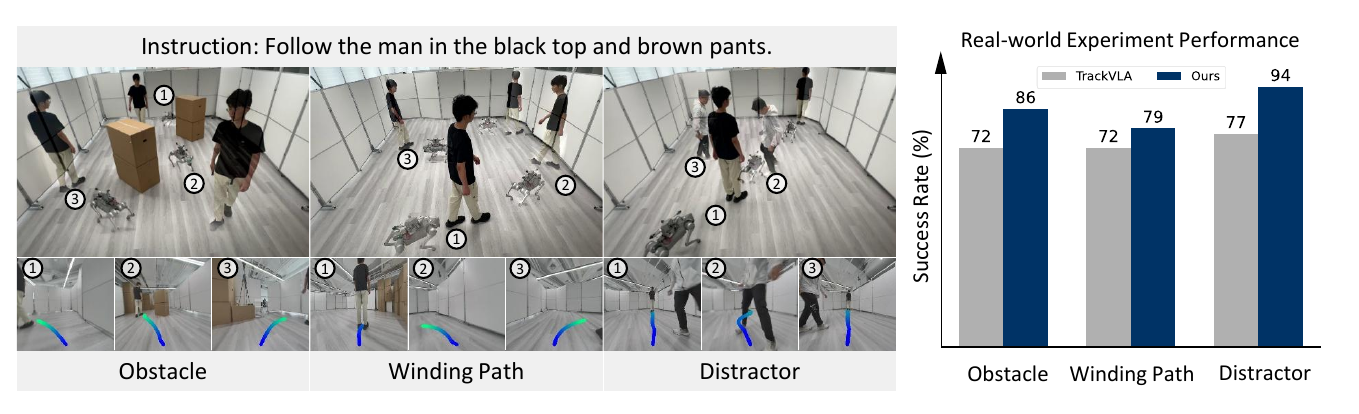}
  \caption{\textbf{Visualizations of the Real World Experiments.} We evaluate \ours on three different tasks: Obstacle, Winding Path, and Distractor, showcasing the tracking performance during target disappearance and occlusion. The bar chart provides a quantitative comparison of success rate between TrackVLA and \ours, highlighting the improved performance of our method.}
  \vspace{-0.5em}
  \label{fig:real_gallery}

\end{figure*}

%% file: table/ablations.tex
\begin{table}[t]
    \caption{\textbf{Ablation Study of Proposed Designs.} We analyze the contributions of individual components on EVT-Bench \texttt{DT split}.}
    \centering
    \begin{tabular}{lccc}
        \toprule
        \multirow{2}{*}{Methods} & \multicolumn{3}{c}{\textit{Distracted Tracking (DT)}} \\
        \cmidrule(lr){2-4}
        & SR $\uparrow$ & TR $\uparrow$ & CR $\downarrow$ \\
        \midrule
        TrackVLA~\cite{wang2025trackvla} & 57.6 & 63.2 & 5.80 \\
        NaVFoM (Four views) & \underline{62.0} & 67.9 & - \\
        \midrule
        \rowcolor{mypurple!20}
        TrackVLA++ (Ours) & \textbf{74.0} & \textbf{73.7} & \textbf{3.51} \\
        \quad w/o Polar-CoT \& TIM & 65.2 & 64.8 & 8.17 \\
        \quad w/o TIM & 71.2 & 69.8 & 4.74 \\
        \quad w TIM (16 tokens) & 74.2 (+0.2) & 73.4 (-0.3) & 3.27 (-0.24) \\
        \bottomrule
    \end{tabular}
    \vspace{-1.5em}
    \label{tab:ablations}
\end{table}

%% file: section/conclusion.tex
\section{Conclusion}
In this paper, we propose \ours, a novel Vision-Language-Action (VLA) model for embodied visual tracking that addresses key limitations of prior approaches by incorporating explicit spatial reasoning and long-horizon target memory. By introducing the polar Chain-of-Thought (Polar-CoT) mechanism and the Target Identification Memory (TIM) module, \ours achieves robust spatiotemporal consistency, effectively handling challenges such as severe occlusions and multiple visually similar distractors.
Extensive experiments demonstrate the effectiveness of \ours, establishing new state-of-the-art performance across simulation benchmarks in both egocentric and multi-camera settings, while also demonstrating remarkable generalization and robustness in real-world scenarios.